\documentclass[letterpaper, 10 pt, conference]{ieeeconf}

\IEEEoverridecommandlockouts                              % This command is only needed if 
\usepackage{graphics} % for pdf, bitmapped graphics files
\usepackage{caption}
\usepackage{ cite}
\usepackage{graphicx}
\usepackage{subfig} 
\usepackage{amssymb}
\usepackage{color}
\usepackage{multirow}
\usepackage[T1]{fontenc}
\usepackage{aecompl}
\usepackage{float}
\title{\LARGE \bf
DeepBlindness: Fast Blindness Map Estimation and Blindness Type Classification for Outdoor Scene from Single Color Image 
}
\author{Jiaxiong Qiu$^{*,1}$, Xinyuan Yu$^{*,2}$, Guoqiang Yang$^{1}$ and Shuaicheng Liu$^{1,\xi}$% <-this % stops a space
	\thanks{This work is supported by Momenta.ai.}% <-this % stops a space
	\thanks{$^{*}$indicates equal contributions.}
	\thanks{$^{\xi}$indicates corresponding author.}
	\thanks{$^{1}$University of Electronic Science and Technology of China}
	\thanks{$^{2}$Momenta.ai}
}
\begin{document}

\maketitle
\thispagestyle{empty}
\pagestyle{empty}

%%%%%%%%%%%%%%%%%%%%%%%%%%%%%%%%%%%%%%%%%%%%%%%%%%%%%%%%%%%%%%%%%%%%%%%%%%%%%%%%
\begin{abstract}

%We propose the first end-to-end convolutional neural network (CNN) architecture for estimating and classifying haze, defocus and motion blur in the outdoor scene. We consider these three types of scenes which limit the clear sight as blindness. Due to lack of almost real dataset to quantify blindness degrees, We construct a novel synthetic blindness dataset named KITTICS, which contains four blindness types(no blindness, haze, defocus and motion blur) of images and their corresponding quantified blindness maps based on KITTI dataset, to train the network. Our network architecture consists of two subnetworks: blindness map estimation and blindness type classification. These subnetworks are supervisely trained in an end-to-end framework. The experimental results on publicly available blur detection dataset show that our method can detect and classify blur effectively and accurately, outperforming other state-of-the-art approaches, and achieves about 108 FPS speed. Our approach can be used for many applications, such as photo editing, blur magnification, dehazing, deblurring, and autonomous driving in challenging scenes.
Outdoor vision robotic systems and autonomous cars suffer from many image-quality issues, particularly haze, defocus blur, and motion blur, which we will define generically as ``blindness issues''. These blindness issues may seriously affect the performance of robotic systems and could lead to unsafe decisions being made. However, existing solutions either focus on one type of blindness only or lack the ability to estimate the degree of blindness accurately. Besides, heavy computation is needed so that these solutions cannot run in real-time on practical systems. In this paper, we provide a method which could simultaneously detect the type of blindness and provide a blindness map indicating to what degree the vision is limited on a pixel-by-pixel basis. Both the blindness type and the estimate of per-pixel blindness are essential for tasks like deblur, dehaze, or the fail-safe functioning of robotic systems. We demonstrate the effectiveness of our approach on the KITTI and CUHK datasets where experiments show that our method outperforms other state-of-the-art approaches, achieving speeds of about 130 frames per second (fps).

\end{abstract}

%%%%%%%%%%%%%%%%%%%%%%%%%%%%%%%%%%%%%%%%%%%%%%%%%%%%%%%%%%%%%%%%%%%%%%%%%%%%%%%%
\section{Introduction}
%Unintentional blindness images captured by cameras in the outdoor scenes can be caused by various reasons. For example, haze, fast moving cars or pedestrians and incorrect focus setting. As an example, Fig \ref{fig1} shows these three types of blindness. Images with degraded quality could make autonomous driving which based on vision unsafe, so it is very important to detect and classify the blindness scenes. Many recent techniques mainly include defocus map estimation \cite{zhuo2011defocus}\cite{shi2015just}, motion blur kernel estimation \cite{whyte2014deblurring}\cite{sun2015learning} to remove motion blur and transmission map estimation \cite{he2010single}\cite{zhang2019joint} to remove haze. But these methods almost handle one type of blindness and other methods for blindness dectection only identify the type without detection. Automatic detection of the partial blindness regions in images and identification of their types are challenging tasks in conputer vision and digital imaging, such as image segmentation \cite{bahrami2013novel}, depth recovery \cite{tang2015depth}, image editing \cite{amin2019hybrid} and self-driving \cite{treml2016speeding}. Blindness detection and classification can also be useful for image deblurring \cite{krishnan2009fast}.

\begin{figure}[h]
	\centering 
	\includegraphics[scale=0.3]{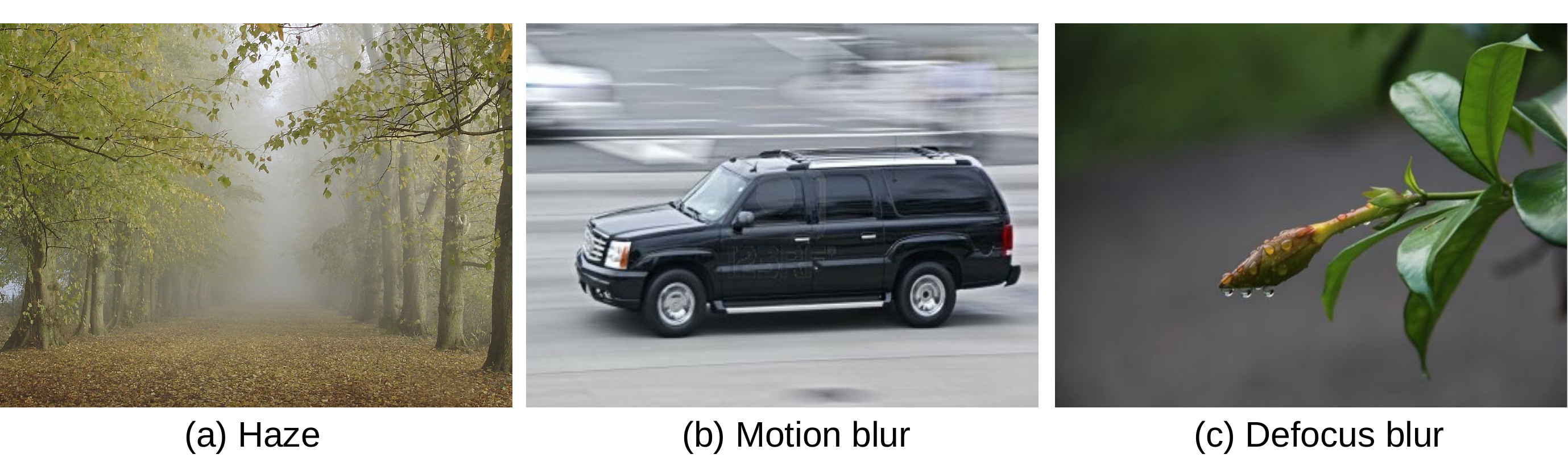}
	\caption{Example of blindness scenes in nature. These scenes may confuse computer vision system.}
	\label{fig1}
\end{figure}

Outdoor robotic systems or autonomous cars frequently suffer from image quality issues. For example, extreme weather conditions, like haze, are a common blindness issue. Also, incorrect focus adjustment may cause defocus blur, and long shutter times often lead to motion blur of dynamic objects.  Fig. \ref{fig1} shows these three types of blindness. Taken together, these blindness issues may seriously affect the reliability of computer vision algorithms, such as object detection or 3D reconstruction. Ultimately, decisions taken by robotic systems due to blindness may be unsafe. Thus, blindness is an important safety factor which should be addressed by stable robotic systems, such as autonomous cars. The ability to classify blindness types intelligently could help the system avoid any serious repercussions of blindness. For example, when motion blur is detected, the system should adjust the shutter time. When defocus blur is found, the focus setting should be set autonomously. And in hazy weather, a foglight should be turned on. Besides, careful estimates of the degree of blindness are also crucial. Deblurring, or dehazing, algorithms will be improved with a per-pixel blindness map. Different decision strategies could also be chosen as a result of estimating the degree of blindness. For instance, it would be totally safe for autonomous cars to drive slowly in hazy weather. 

Many blindness-related tasks are widely studied, for example, defocus map estimation, motion blur kernel estimation and haze-related transmission map estimation. Lots of blur detection approaches propose to design hand-crafted features to differentiate clear-sight and business regions. These approaches use textural information or edge sharpness to estimate the blur amount initially and then apply it to the whole image. It is difficult to provide an accurate blur estimation map by simply using a propagation-based method.  Zhang \textit{et al.} \cite{zhang2015fast} proposed an efficient method of recovering a robust estimate of motion blur kernel by simplifying the wavelet analysis of a blurry image at low speed. Chen \textit{et al.} \cite{chen2018haze} introduced a haze density framework for estimating scattering coefficients of iso-depth regions, though with less accurate results. In recent years, many convolutional neural network (CNN) approaches have used high-level features of patches of blurry images to reconstruct a blur map. Ma \textit{et al.} \cite{ma2018deep} proposed a real-time method by only using the low-level features to get the blur map. However, their results suffered from an ambiguous blindness definition and usually assumed that the blindness type was already known. 

Most of the above methods have obvious shortcomings. Blur detection only provides a discrete binary blindness map and highly depends on human annotation. The ambiguous definition of the blindness region has brought subjective bias into the annotation procedure. Besides, a binary blindness map is usually not enough for tasks like dehazing or deblurring. Lack of datasets is also a critical issue. Another barrier to using any of these approaches in a practical application is that the blindness type is usually unknown and hard to predict. In fact, many deblur methods have to assume that the blindness type is given already. In addition, the complexity of the method and the running time also have to be considered. 

Ideally, we believe a method should be able to classify the blindness type and provide an accurate per-pixel estimate of the blindness map (including being able to indicate which type the vision is limited by) in real time.

In this paper, we propose an efficient and flexible CNN-based architecture which can accomplish the above requirements. As distinct from the previous methods, our network includes three subnetworks, which are respectively about haze, defocus blur and motion blur. Each subnetwork gives a per-pixel blindness map and a confidence level representing whether the image has an individual blindness issue. Afterwards, we integrate these outputs into a single blindness map according to the identified blindness type. To train the network, we need a dataset including all three blindness types. A trusted overall blindness map is essential as well. To the best of our knowledge, there are no publicly available datasets that meet our requirement. Thus, we had to generate a synthetic dataset containing the three different blindness types. We also provide degree of blindness evaluation metrics which indicate the overall truth of the per-pixel blindness map without human annotation. The experimental results show that our method is faster than current state-of-the-art methods and achieves comparable accuracy for blindness detection and classification.

Here is a summary of the three main contributions from our work: 

$\bullet$ We provide both a scientific method to quantize the degree of vision-limiting caused by the three different blindness issues, and objective ground-truth metrics. Thus, our CNN network can be trained without ambiguity or time-consuming human annotation.

$\bullet$ We generate a synthetic dataset based on real, autonomous driving scenes. Our dataset includes four specific types of image: clear, haze, defocus blur, and motion blur. From these, we derive an accurate ground-truth blindness map which can be used to train the model and hence transfer successfully to real scenes.  

$\bullet$ We propose a fast, efficient and flexible CNN-based architecture for blindness detection and classification using a simple training strategy. This achieves more than 200 fps speed on blindness map estimation and about 130 fps in total.

\section{Related Work}

\subsection{Blur Detection and Classification}
For blur detection and classification, many previous approaches used hand-crafted features, such as edge sharpness information. Liu \textsl{et al.} \cite{liu2008image} developed some blur features and used them to classify two types of blur images. Chakrabarti \textsl{et al.} \cite{chakrabarti2010analyzing} analyzed directional blur by local Fourier transform. Su \textsl{et al.} \cite{su2011blurred} proposed to detect blurred image regions by examining singular value information, then determined the blur types with certain alpha channel constraint. Shi \textsl{et al.} \cite{shi2014discriminative} studied several blur feature representations about image gradient, Fourier domain, and data-driven local filters. They also proposed a dataset which contains 1,000 images with labeled binary map for motion blur and defocus blur detection and label of blur type. 
Tang \textsl{et al.} \cite{tang2016spectral} presented a blur metric to get a coarse blur map and then exploited the intrinsic relevance of similar neighbor image regions to refine the coarse blur map. Golestaneh \textsl{et al.} \cite{golestaneh2017spatially} computed blur maps by a high-frequency multi-scale fusion and sorted transform of gradient magnitudes. These methods heavily depends on the accuracy of hand-crafted blur features. However, accurate features are hard to achieve. 

Nowadays, with the fast development of deep learning, some CNN-based approaches sprang up to detect blur images and classify them, and achieved more efficient results. Huang \textsl{et al.} \cite{huang2018multiscale} proposed to learn discriminative blur features learned by CNN, they designed a 6-layer CNN model and fused multi-scale blur maps to get the final blur map. Some of their results were incorrectly aligned with source object boundaries. Ma \textsl{et al.} \cite{ma2018deep} designed a fully convolutional network (FCN) for blur detection only, their blur maps lack clear boundaries of objects, because they did not consider low-level features of input images. Kim \textsl{et al.} \cite{kim2018defocus} proposed a deep encoder-decoder network to detect and classify defocus and motion blur with complicated training strategy. They also constructed a synthetic dataset which simply combined both of these two types of blurry objects. Their results were unstable in some scenes, where these results usually made no sense. Besides, these architectures cost much time to detect the blur regions in blurry images, some of them have the assumption that the blur type is already known and others need complicate training process.  

\subsection{Defocus Map Labeling}
There are some approaches proposed to densely label defocus map. Andr\`es \textsl{et al.} \cite{d2016non} constructed a dataset which labeled with the radius of point-spread-function (PSF) per pixel. This method was inefficient to deal with pixels around boundaries of depth map. Zhang \textsl{et al.} \cite{zhang2018learning} introduced a dataset labeling each pixel of blurry images by four degrees of blur: high, medium, low, and no blur. They neither gave fine blur amount nor dataset annotated with a definite rules. Lee \textsl{et al.} \cite{lee2019deep} synthesised a defous blur dataset, but it was almost based on virtual datasets and unsuitable for data-driven CNN-based approaches. 

%\subsection{Transmission Map for Haze Estimation}
%Transmission map estimation was a kind of convenient media for dehazing. There are also some methods introduced to solve this problem. Guo \textsl{et al.} \cite{guo2014foggy} designed a serial steps which consist of image segmentation, initial map estimation based on MRF, and refined map estimation using bilateral filter for transmission map estimation. Whereas, their work was time-consuming and could be affected by different objects’ colors. Wang \textsl{et al.} \cite{wang2016transmission} proposed a hybrid of a recurrent fuzzy cerebellar model articulation controller (RFCMAC) to estimate the transmission map, this method was a bit complex and hard to achieve high accuracy. Zhang \textsl{et al.} \cite{zhang2019joint} constructed a dataset which took transmission maps as ground-truth based on a indoor dataset. They used depth maps to generate transmission maps by a simple equation. However, their dataset did not include outdoor scenes.

Inspired by the success of above approaches in blur detection, blur classification and defocus blur and haze dataset generation method, we proposed a high-efficiency method for constructing  blindness dataset named KITTICS, which covers outdoor scenes. In the mean time, we provide a real-time blindness detection and blur type classification system. 

\begin{figure*}[h]
	\vspace{0.1cm}
	\centering 
	\includegraphics[scale=0.55]{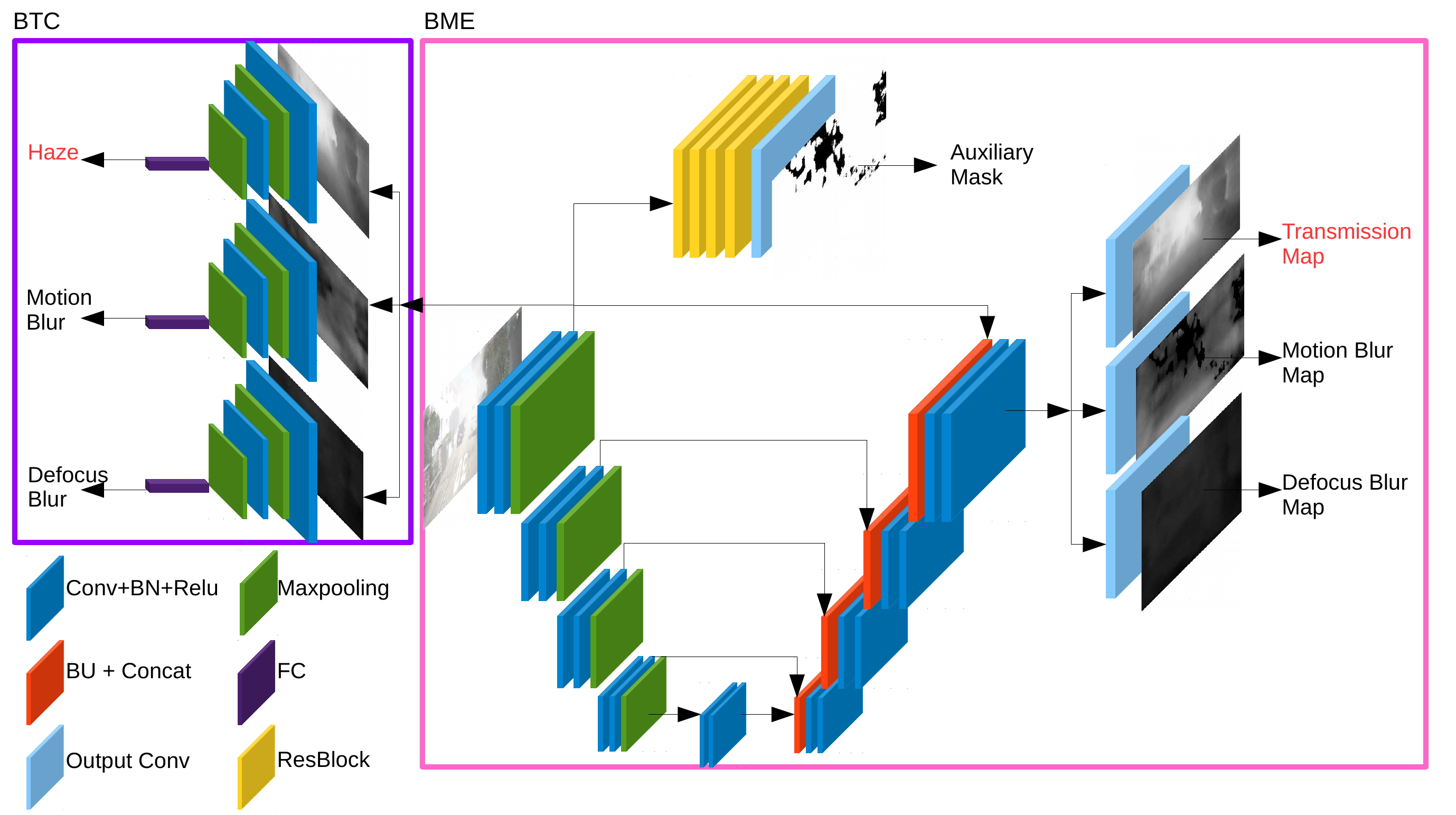}
	\caption{Our blindness map estimation and type classification network architecture. We use a hazy image as example, the left red text represents the blindness type with the highest probability which our \textbf{\textit{BTC}} outputs, the right red text expresses the blindness map produced by our \textbf{\textit{BME}}.}
	\label{fig_ov}
	\vspace{-0.6cm}
\end{figure*}

\section{Dataset Construction}

We concentrate on real and challenging autonomous driving scenes which frequently suffer from blindness issues. The KITTI benchmark provides a clear sequence of images with LiDAR data providing depth information within the images. Such information is essential for generating blindness images.

\subsection{KITTI Dataset}
Uhrig \textit{et al.} \cite{uhrig2017sparsity} developed a dataset from the KITTI benchmark comprising 93 thousand depth-annotated RGB images. They projected the raw LiDAR data to the aligned RGB images to get a sparse LiDAR depth map. They also raised a depth completion competition by using this dataset. Qiu \textit{et al.} \cite{Qiu_2019_CVPR} proposed a state-of-the-art system which took as input sparse LiDAR depth maps and RGB images and generated robust depth maps.  

To construct pairs of hazy images and transmission maps, and pairs of defocus images and defocus maps, we used the model of  \cite{Qiu_2019_CVPR} to get to get an accurate dense depth map.  

\subsection{Hazy Image Generation}

As proposed by \cite{koschmieder1924theorie} and \cite{narasimhan2002vision}, the atmospheric scattering model is widely used to describe the formation of the hazy image. Hence, the hazy effect can be mathematically formulated as: 
\begin{equation}\label{q1}
\textbf{I}\textup{(x)} = \textbf{J}\textup{(x)}\textit{t}\textup{(x)} + \textbf{A}\textup{(x)}\textup{(1 - \textit{t}(x))}
\end{equation}
where \textup{x} is the pixel coordinates, \textbf{I} stands for the observed hazy image, \textbf{J} is the true scene radiance which means the clear image before degradation, \textit{t} is the transmission map which is a distance-dependent factor and represents the relative portion of light that manages to survive the entire path between the observer and a surface point in the scene without being scatted, and \textbf{A} is the global atmospheric light. We use this equation \ref{q1} to generate hazy images. 

We calculate the \textbf{A} by the method proposed in \cite{he2010single}. With dense depth prepared by \cite{Qiu_2019_CVPR}, the transmission map can be expressed as:
\begin{equation}\label{q2}
\textit{t}\textup{(x)} = e^{-\beta\textit{d}\textup{(x)}}
\end{equation}
where the $\beta$ is the attenuation coefficient of the atmosphere and \textit{d} represents the depth of the scene. The transmission map can express the regions with clear sight, so the reverse map of the transmission map can represent the haze amount. The scale of the transmission map is between 0.0 and 1.0, so we set 1.0-\textit{t} as the ground-truth of haze amount.    

\subsection{Defocused Image Generation}

We follow the approach proposed by \cite{carvalho2018deep} to generate defocused images, which is based on two equations:
\begin{small} 
\begin{equation}\label{q3}
\widehat{L} = \sum_k[(A_kL + A^*_kL^*_k)*h(k)]M_k
\vspace{-0.5cm}
\end{equation}
\begin{equation}\label{q4}
M_k = \prod_{k'=k+1}^K(1-A_{k'}*h(k'))
\end{equation}
\end{small}
where $\hat{L}$ is the defocus blurred image which is the sum of a certain number of blurred images multiplied by masks taking into account occlusion of foreground objects and local object depth. $k$ is the depth of the scene, $h$ is the defocus blur function, $L$ is the all-in-focus image, $A_k$ is a binary mask related to object at depth $k$ and represents location of the depth. We choose disk function with varied diameter according to depth. The diameter map can represent the defocus blur amount, so we take the diameter as the ground-truth of defocus blur amount.

\subsection{Motion Blurred Image Generation}
Inspired of the method \cite{brooks2019motionblur}, we use an efficient and traditional frame interpolation method \cite{meyer2015phase} to synthesize motion-blurred images. To construct our dataset, we firstly extract sets of adjacent pairs of continuous frames and then apply \cite{meyer2015phase} to interpolate three frames between each pair, making five frames in total. Then, we synthesize a motion blur image by averaged sum of the five frames. 

\begin{figure}[thpb]
	\vspace{-0.5cm}
	\centering 
 	\includegraphics[scale=0.34]{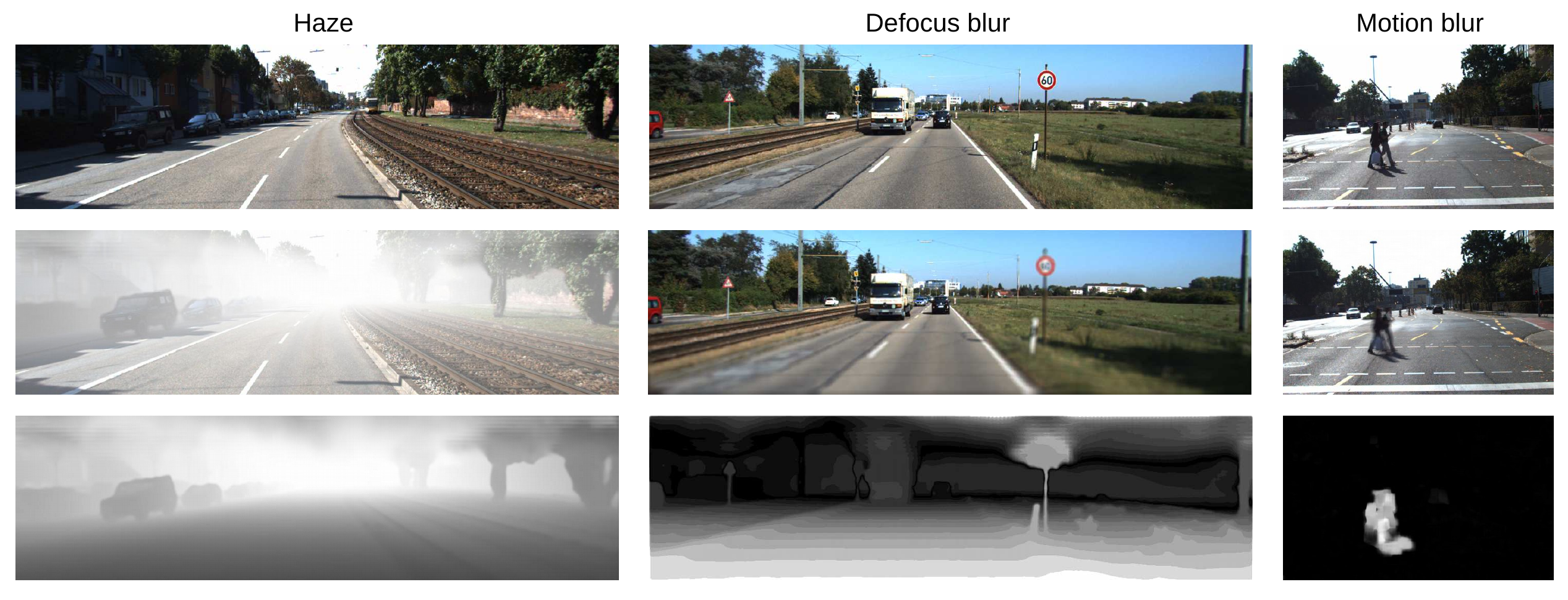}
	\caption{Example of our KITTICS dataset. The first row: origin images in the KTIITI dataset \cite{uhrig2017sparsity}. The middle row: synthetic images with our method. The last row: ground-truth maps.}
	\label{fig2}
	\vspace{-0.5cm}
\end{figure}

The motion blur amount is estimated by the dense optical flow magnitude, which is calculated from the \cite{farneback2003two}. We introduce the magnitude of the dense optical flow as the ground-truth of motion blur. Fig. \ref{fig2} shows some examples of our blindness dataset, including 39,991 images of the four different blindness types, which are clear, haze, defocus blur and motion blur, and their corresponding pixel-level ground-truth of blindness amount.

As shown in table \ref{table1}, our dataset provides larger amount of images, and more blindness types, than all other existed datasets. Images used in other datasets are mostly coarsely-annotated or unreal. Our KITTICS dataset provides a high reliable training and evaluation platform for outdoor blur analysis in autonomous safe-driving research. It is an expandable dataset, which also can be used in video and image deblurring. Table \ref{table2} explains the composition of KITTICS. 
~\\
\begin{table}[h] 
	\vspace{-0.5cm}
	\begin{center}
		\resizebox{87mm}{13mm}{
		\newcommand{\tabincell}[2]{\begin{tabular}{@{}#1@{}}#2\end{tabular}}
		\begin{tabular}{|c|c|c|c|c|}
			\hline
			Dataset & CUHK\cite{shi2014discriminative} & SmartBlur\cite{zhang2018learning} & SYNDOF\cite{lee2019deep} & \textbf{KITTICS}\\
			\hline
			\# of Images & 1000 & 10,000 & 8231 & \textbf{39,991}\\
			\hline
			Blindness Type & 1,2 & 1,2 & 2 & \textbf{1,2,3}\\
			\hline
			Blindness Amount & \tabincell{c}{Pixel-wise \\  binary} &  \tabincell{c}{Pixel-wise \\ multi-level} & Depth-wise & \textbf{\tabincell{c}{Depth-wise \\ or pixel-wise}}\\
			\hline
			Automation &  $\times$ & $\times$ & $\checkmark$ & \textbf{$\checkmark$}\\
			\hline
			Image Source & Natural & Natural & \tabincell{c}{Sythetic \\ and virtual} & \textbf{\tabincell{c}{Synthetic \\ and natural}}\\
			\hline
		\end{tabular}
		}
	\caption{Comparison of blindness image datasets. 1,2,3 represent blindness types and indicate motion blur, defocus and haze respectively.}
	\label{table1}
	\end{center}
	\vspace{-0.5cm}
\end{table}

\begin{table}[h]
	\begin{center}
		\resizebox{86mm}{8mm}{
			\begin{tabular}{|c|c|c|c|c|c|}
				\hline
				Blindness type & No blindness & Haze & Defocus blur& Motion blur & Total \\
				\hline
				Training & 9,799 & 9,798 & 9,799 & 9,799 & 39,195\\
				\hline 
				Testing  & 198 & 200 & 199 & 199 & 796\\
			    \hline 
			    Total & 9,997 & 9,998 & 9,998 & 9,998 & 39,991 \\
			    \hline
			\end{tabular}
	     }
	\caption{The image amount of split the KITTICS with different types.}
	\label{table2}
	\end{center}
	\vspace{-1cm}
\end{table}

\section{Blindness Map Estimation and Blindness Type Classification}

\subsection{Network Architecture}
As shown in the overview Fig. \ref{fig_ov}, we design a novel and lightweight network architecture, which can simultaneously estimate a blindness map and classify the blindness type. The network architecture consists of two components: blindness map estimation (\textbf{\textit{BME}}) and blindness type classification (\textbf{\textit{BTC}}). Inspired by the U-Net architecture \cite{ronneberger2015u}, our \textbf{\textit{BME}} is constructed upon an encoder-decoder framework with long skip-connections to use both low-level and high-level features. The encoder part of the \textbf{\textit{BME}} consists of four max-pooling layers and eight convolution layers. 

In order to achieve a high speed and reduce the training time and memory needed, the decoder part of the \textbf{\textit{BME}} consists of four bilinear upsampling and eight convolution layers to generate the output with the same size of the input. 

All the convolution layers are followed by batch-normalization and relu operations following. At the end of the \textbf{\textit{BME}}, we apply three convolution layers for producing three blindness maps simultaneously. By analyzing the motion blur map, we find that it has a close correlation to the object boundary. While the motion blur map has different value inside the boundary, it is mostly close to zero outside. On the basis of this, we design a subnetwork which consisting of four resblock \cite{he2016deep} layers and one convolution layer to learn, and an auxiliary mask to locate the motion blur region. Given the motion blur region, the motion blur degree can be learned more specifically. The final motion blur map is the product of the predicted motion blur map and the auxiliary mask. We use three VGG-4 \cite{simonyan2014very} networks in parallel to build our \textbf{\textit{BTC}}. Each VGG-4 network consists of two convolution layers, two max-pooling layers and one fully-convolution layer. 

We concatenate the image encoder and the estimated blindness map. Then, we use the concatenated result as the input of \textbf{\textit{BTC}}, which specifically outputs the probability of three types of blindness. All the outputs of our network are joint with a sigmoid operation to ensure their value between 0.0 and 1.0. Our network finally generates six outputs, three classification probabilities and three blindness maps, which can be efficiently fused for a rich source of information. 

\subsection{Loss Functions}

We use regular L1 losses ($L_1$) for blindness map estimation $L_{BM}$. To train our network, binary-cross-entropy losses ($L_{BCE}$) are used for both blindness type classification $L_{BT}$ and auxiliary mask estimation $L_{AM}$. These three loss functions are defined as:

\begin{footnotesize} 
\begin{equation}\label{q7}
L_{BM}=
\left\{ 
\begin{array}{lr}
\sum_{c_t}L_1(b_{c_t},\hat{\textbf{0}}), c=-1 , c_t=0,1,2 \\
\sum_cL_1(b_c,\hat{b}_c), c=0,1,2
\end{array} 
\right.
\end{equation}
\vspace{-0.1cm}
\begin{equation}\label{q8}
L_{BT}=
\left\{ 
\begin{array}{lr}
\sum_{c_t}L_{BCE}(x_{c_t},\hat{\textbf{0}}), c=-1 , c_t=0,1,2 \\
\sum_c[L_{BCE}(x_c,\hat{\textbf{1}}_c)+\sum_{\overline{c}}L_{BCE}(x_{\overline{c}},\hat{\textbf{0}}_{\overline{c}})]. c=0,1,2
\end{array} 
\right.
\end{equation}
\vspace{-0.1cm}
\begin{equation}\label{q9}
L_{AM}=
L_{BCE}(m_c,\hat{m}_c)+\sum_{\overline{c}}L_{BCE}(m_{\overline{c}},\hat{\textbf{0}}_{\overline{c}}). c=1
\end{equation}
\vspace{-0.3cm}
\end{footnotesize} 

where $b$ is the estimated blindness map by our network, and $\hat{b}$ is the ground-truth blindness map. $c$ is current blindness type (-1,0,1 and 2 are representing no blur, haze, motion blur and defocus blur respectively), $c_t$ is all blindness types and $\overline{c}$ is the different blindness types from $c$. $x$ is the predicted probability of blindness type. $\hat{\textbf{0}}$ is a tensor filled with zero, and $\hat{\textbf{1}}$ is a tensor filled with one and they are as the same size as $x$. $m$ is the predicted auxiliary mask for motion blur and $\hat{m}$ is the ground-truth of it. Training can be done by minimizing the loss function defined as:
\begin{equation}\label{q10}
L = L_{BM} + \lambda_1 L_{BT} + \lambda_2 L_{AM}
\end{equation}
where $\lambda$ is a weight parameter of $L_BT$. 

\subsection{Implementation Details}
In order to speed up training and testing and reduce GPU memory cost, we resize the input image to 256 $\times$ 256. Then we perform some normal data augmentation operations. 

Our network is implemented with Pytorch and trained on eight GeForce GTX 1080 Ti GPUs. We use Xavier initialization \cite{glorot2010understanding} to initialize the whole network and the Adam optimizer \cite{kingma2014adam} as the optimizer during training. We set 0.001 as the initial learning rate. 

Our training strategy is very simple and ordinary. When being compared on CUHK dataset, we follow \cite{kim2018defocus} and split 1,000 images into two parts which contain 800 images for training and 200 images which are only used for quantitative evaluation. We take 100 epochs for training on our KITTICS dataset and 800 epochs on the CUHK dataset.

When evaluating on the CUHK dataset, we propose a fusion strategy to generate the final blindness map:
\begin{equation}\label{q11}
b_{final} = b_{mb}p_{mb} + b_{db}p_{db} + \beta m
\end{equation}
where $b_{final}$ is the fused blindness map. $b_{mb}$ and $b_{db}$ are the estimated motion blur map and defocus blur map respectively by our network. $p_{mb}$ and $p_{db}$ are the predicted probability of motion blur and defocus blur respectively. $\beta$ is the weight of auxiliary mask $m$. We set $\beta$ = 0.1 in evaluation. As the ground-truth of the CUHK dataset contains only binary maps and our network produces three labels of each pixel, we convert $b_{final}$ into binary blindness maps by following the method of \cite{park2017unified}. The threshold $\tau$ for binarization is defined as:
\begin{equation}\label{q12}
\tau = \alpha v_{max} + (1-\alpha)v_{min}
\end{equation}
where $v_{max}$ is the maximum values in the estimated blindness map, $v_{min}$ is the minimum values in the estimated blindness map and $\alpha$ = 0.455.

\section{Experimental Results}

\subsection{Evaluation on the CUHK Dataset}

Among 200 images from the CUHK dataset for quantitative evaluations, 60 images are motion blurred and the other 140 images are defocus blurred. We adopt the four metrics that Kim \textit{et al.}  \cite{kim2018defocus} used and speed evaluation, to evaluate the performance of our network and compare with other methods. Although other approaches did not predict the classification probability of the motion blur and the defocus blur, we also evaluate the classification accuracy. The three metrics used to evaluate the predicted binary blindness maps are accuracy, mean intersection of union (mIoU) and F-measure. They are defined as:

\begin{small}
\begin{equation}\label{q13}
Accuracy = \frac{1}{N}\sum_p(1-|b_p-\hat{b}_p|)
\end{equation}
\begin{equation}\label{q14}
mIoU = \frac{1}{N_c} \sum_c(\frac{b_c \cap \hat{b}_c}{b_c \cup  \hat{b}_c})
\end{equation}
\begin{equation}\label{q15}
precision = \frac{\sum_p b_p\hat{b}}{\sum_p b_p}, recall = \frac{\sum_p b_p\hat{b}}{\sum_p \hat{b}_p}
\end{equation}
\begin{equation}\label{q16}
F_{\beta}\rule[2pt]{0.1cm}{0.05em}measure = \frac{(1+\beta)precision \times recall}{\beta^2precision + recall}
\end{equation}
\end{small}

\begin{table}[h]
	\begin{center}
		\vspace{0.2cm}
		\newcommand{\tabincell}[2]{\begin{tabular}{@{}#1@{}}#2\end{tabular}}
		\resizebox{86mm}{10mm}{
			\begin{tabular}{|c|c|c|c|c|c|c|c|c|}
				\hline
				& \tabincell{c}{Pre-trained \\ model} & \tabincell{c}{Trained with \\ synthetic data} & Accuracy$\uparrow$ & $\sigma^2$ of accuracy$\downarrow$  & F-measure$\uparrow$ & mIoU$\uparrow$ & Speed$\uparrow$ & \tabincell{c}{Classification \\ accuracy}$\uparrow$\\
				\hline
				Shi \textit{et al.} \cite{shi2014discriminative} & $\times$ & $\times$ & 0.7423 & 0.0220  & 0.8055 & 0.5617 & 0.002 &-\\
				\hline
				Tang \textit{et al.} \cite{tang2016spectral} & $\times$ & $\times$ & 0.7701 & 0.0322 & 0.8295 & 0.6040 & 0.976 &-\\
				\hline
				Golestaneh \textit{et al.} \cite{golestaneh2017spatially} & $\times$ & $\times$ & 0.8010 & 0.0227  & 0.8417 & 0.6349 & 0.010 &-\\
				\hline
				Huang \textit{et al.} \cite{huang2018multiscale} & $\times$ & $\times$ & 0.7949 & 0.0221  & 0.8485 & 0.6318 & 0.022 &-\\
				\hline
				Ma \textit{et al.} \cite{ma2018deep} & $\checkmark$(VGG) & $\checkmark$ & \textcolor[rgb]{0,0,1}{0.8827} & \textcolor[rgb]{1,0,0}{0.0114}  & \textcolor[rgb]{0,0,1}{0.8883} & \textcolor[rgb]{0,0,1}{0.7578} & \textcolor[rgb]{0,1,0}{37.037} &-\\
				\hline
				Kim \textit{et al.} \cite{kim2018defocus}& $\checkmark$(VGG) & $\checkmark$ & \textcolor[rgb]{1,0,0}{0.9034} & \textcolor[rgb]{0,0,1}{0.0139}  & \textcolor[rgb]{1,0,0}{0.8961} & \textcolor[rgb]{1,0,0}{0.8047} & \textcolor[rgb]{0,0,1}{2.893} &-\\
				\hline
				Ours & $\checkmark$(KITTICS) & $\times$ & \textcolor[rgb]{0,1,0}{0.8937} & \textcolor[rgb]{0,1,0}{0.0133} & \textcolor[rgb]{0,1,0}{0.8895} & \textcolor[rgb]{0,1,0}{0.7737} & \textcolor[rgb]{1,0,0}{129.282} & 0.935\\
				\hline
				Ours-KITTICS & $\times$ & $\checkmark$ (only)& 0.6909 & 0.0318  & 0.7637 & 0.4322 & 129.282 & 0.455\\
				\hline
			\end{tabular}
		}
		\caption{Quantitative comparison of blindness detection performance with other blur detection approaches and blindness type classification accuracy of our network predicted on the CUHK test dataset. $\uparrow$ and $\downarrow$ denote that larger and smaller is better, respectively. The best three results are marked in red, blue, and green respectively. The Ours-KITTICS means that our network trained on our KITTICS dataset only.}  
		\label{table3}
	\end{center}
	\vspace{-0.4cm}
\end{table}

\begin{figure}[h]
	\vspace{-0.4cm}
	\centering
	\subfloat[Precision-recall curves]{
		\includegraphics[scale=0.2]{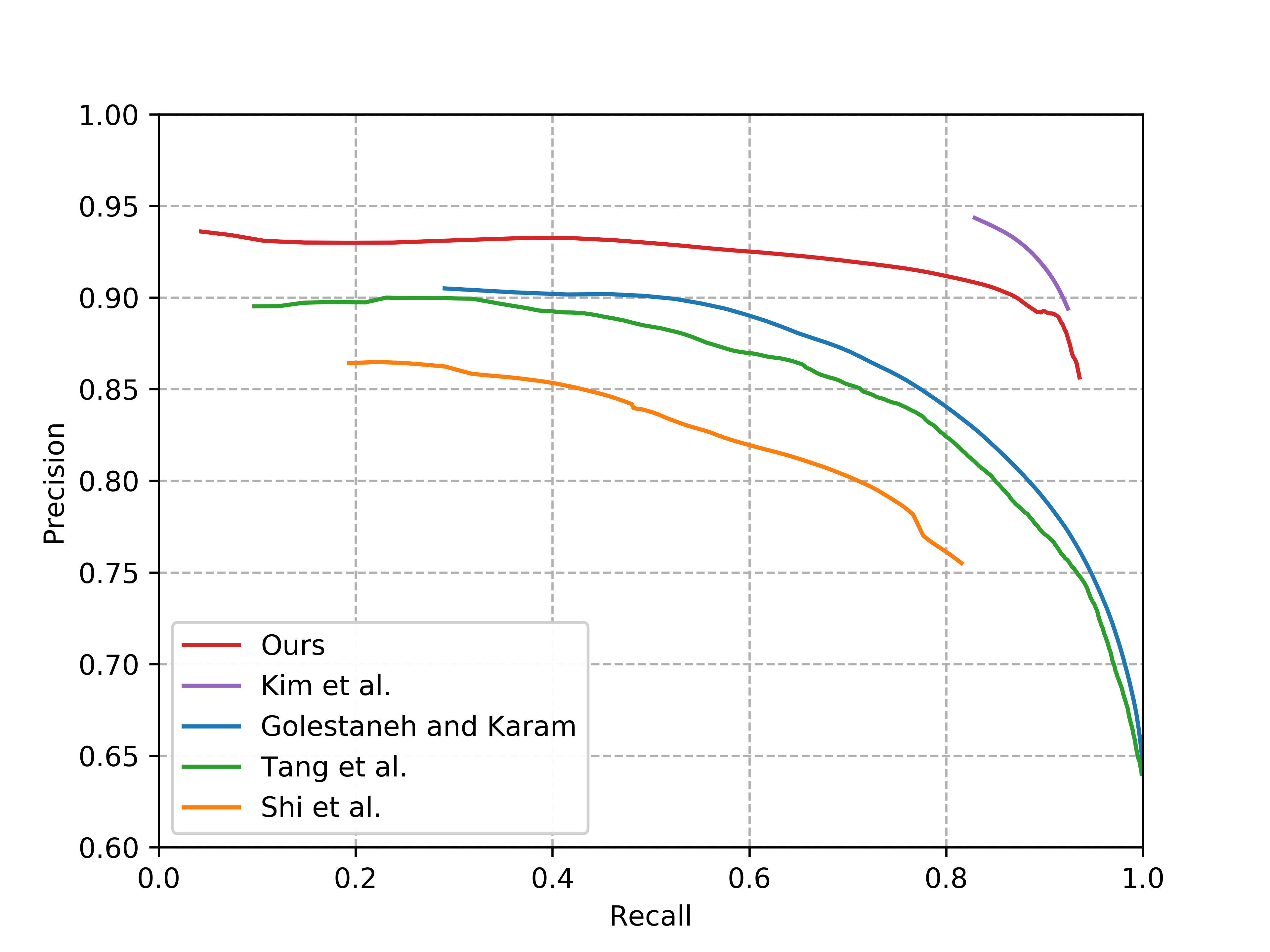}
	}
	\subfloat[Mean precision (mp), mean recall (mr), and speed comparisons ]{
		\includegraphics[scale=0.25]{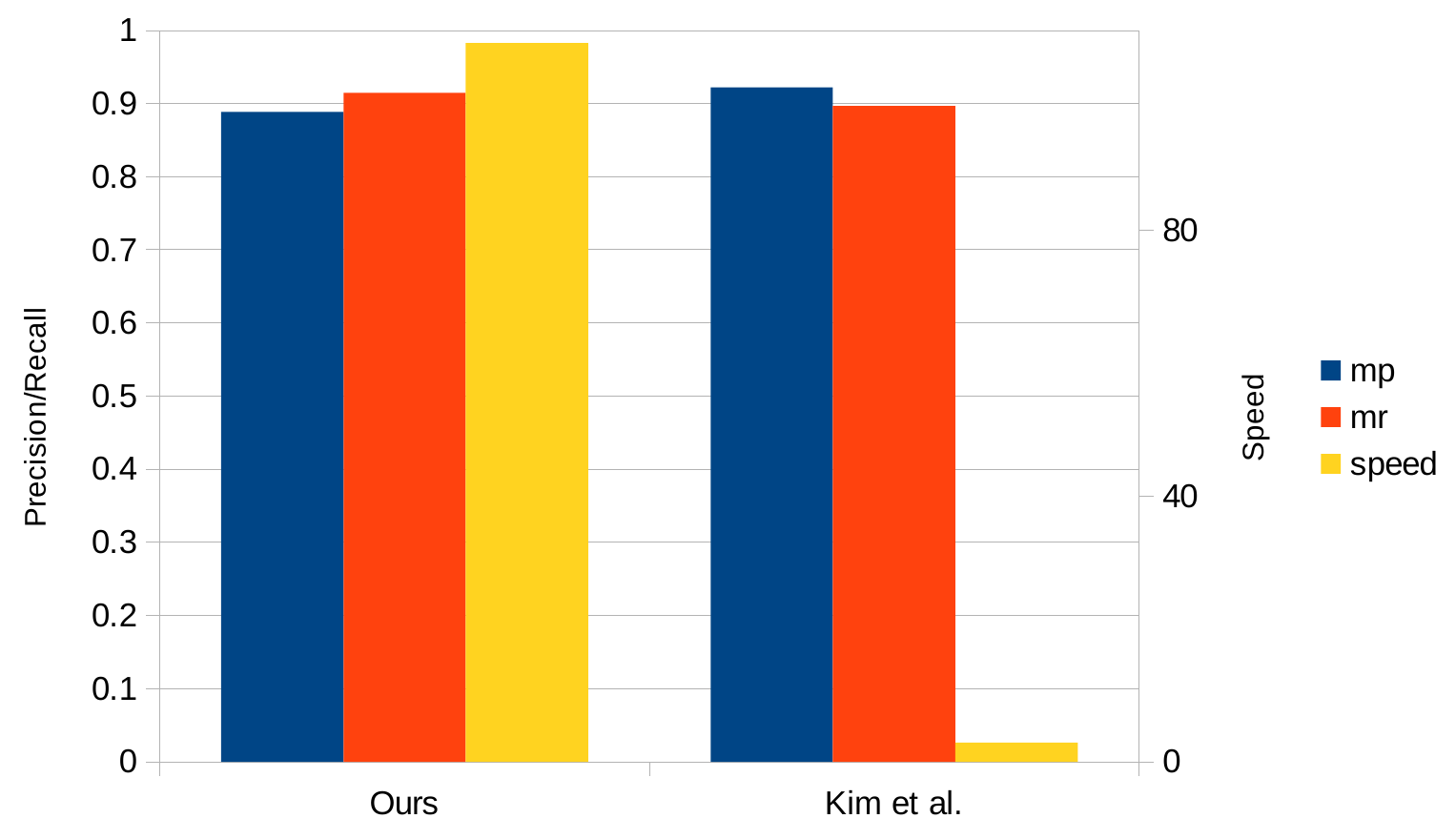}
	}
	\caption{Two comparable graphs for blindness detection performance from Table \ref{table3}. Our network is pre-trained on our KITTICS dataset. Please note that our network is trained without any other complex training strategy.}
	\label{fig3}
	\vspace{-0.7cm}
\end{figure}

where $N_c$ is the number of blindness types in a blindness map. Because the ground-truth only contains clear sight and blind sight, the $N_c$ is two. $\beta$ is a parameter of the F-measure. In order to make precision and recall have the same importance, we set the $\beta$ = 1. Kim \textit{et al.}  \cite{kim2018defocus} introduced the variance ($\sigma^2$) of the accuracy of the whole test images to represent the stability of each method. In addition to these metrics for measuring the quantity of blindness maps, we also compute the speed as the reciprocal of the average processing time. The units of speed is frame-per-second (FPS).

\begin{figure}[h]
	\vspace{0.2cm}
	\centering
	\includegraphics[scale=0.3]{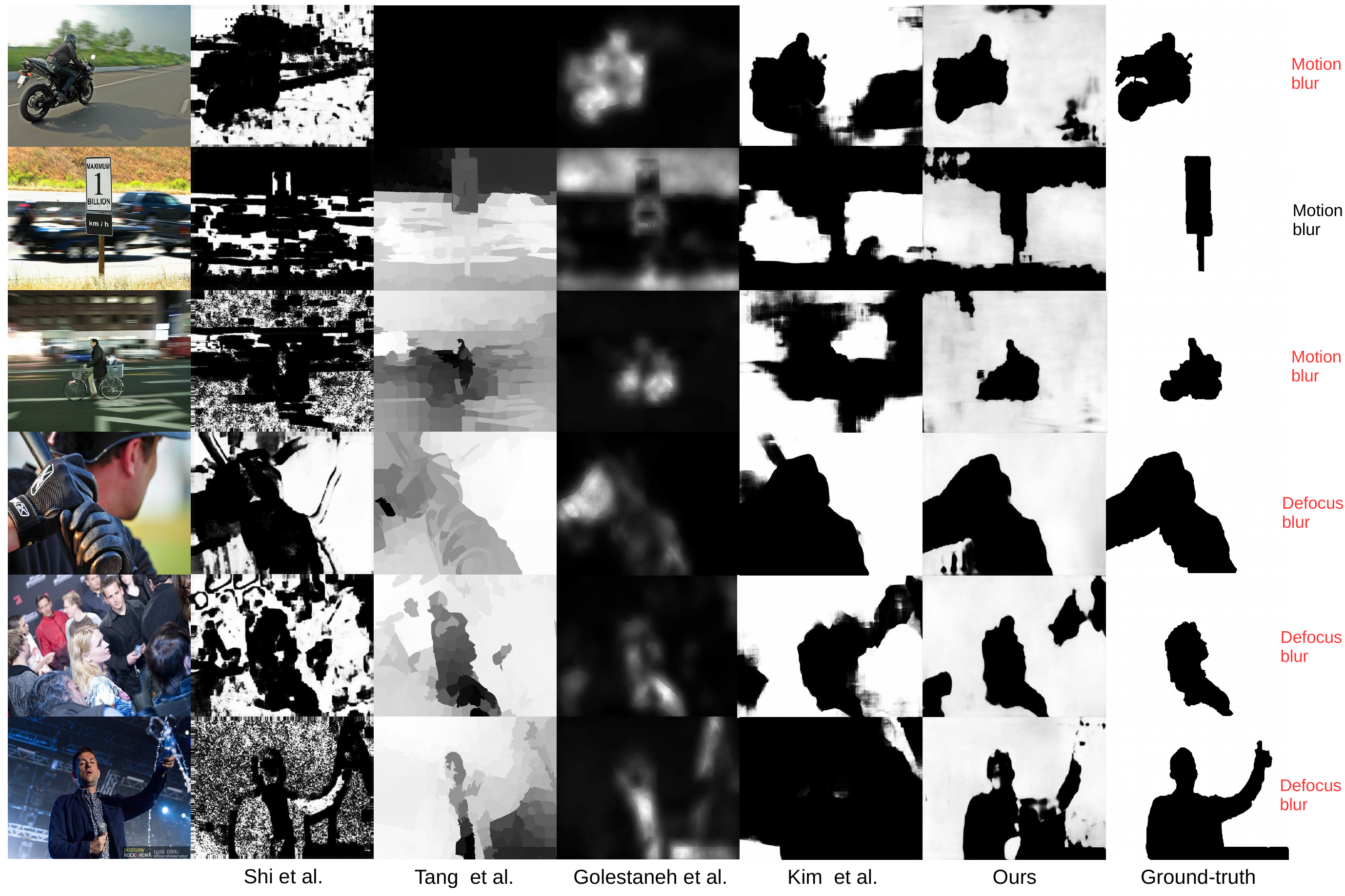}
	\caption{The blindness map estimation results of our method compared with other blindness map estimation approaches. White or black colors in the blur map represents the blindness regions or clear regions, respectively. The last column shows the blindness type classification results of our method and the red text means correct prediction. Only the second case failed. However, the annotation of the second case is ambiguous.}
	\label{fig4}
	\vspace{-0.5cm}
\end{figure}

\begin{figure}[h]
	\centering
	\includegraphics[scale=0.34]{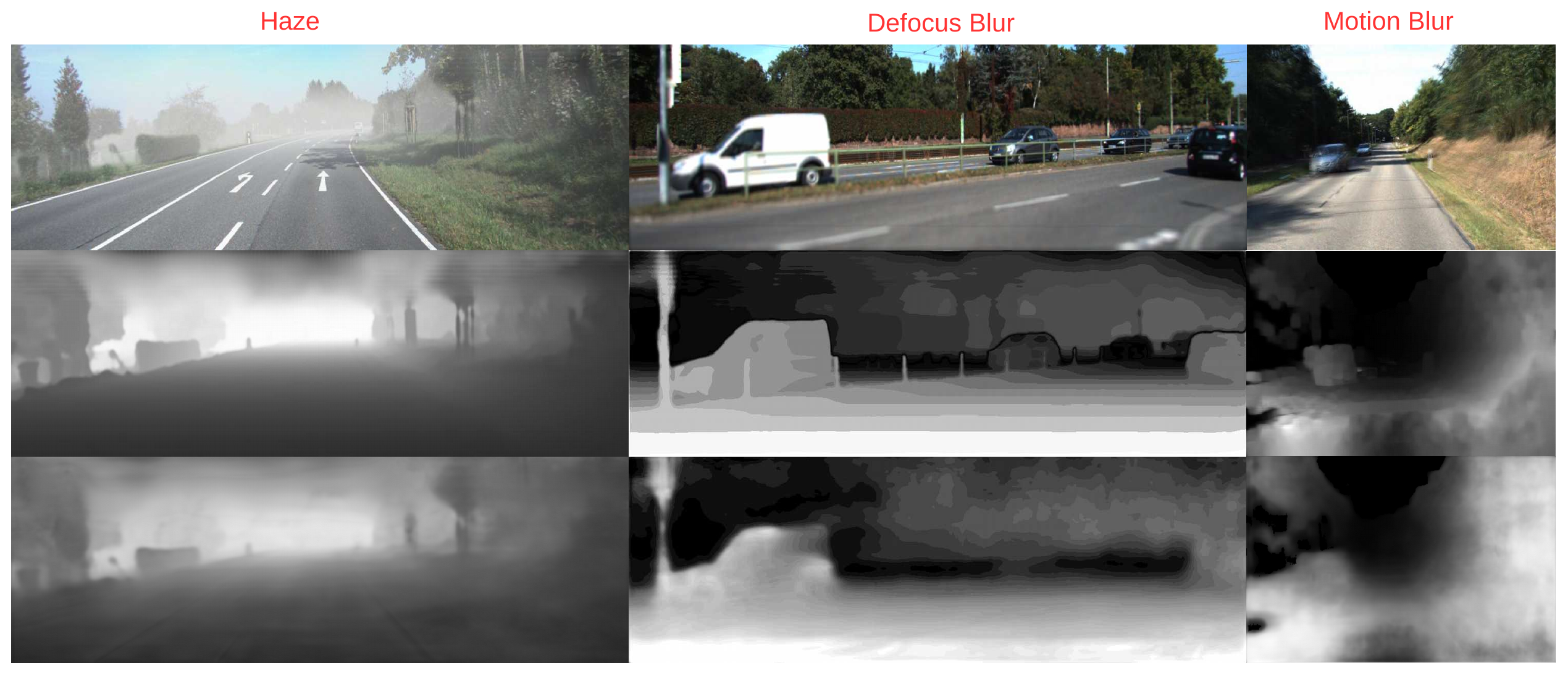}
	\caption{Example of our results on our KITTICS dataset. The red texts: blindness type predictions. The first row: input images. The middle row: the ground-truth maps. The last row: the predicted blindness maps.}
	\label{fig5}
	\vspace{-0.5cm}
\end{figure}

As shown in Table \ref{table3}, we compare our approach with other state-of-the-art blur detection methods and we directly use the value of metrics from \cite{kim2018defocus}. About the \cite{kim2018defocus}, we use the code and model of it to compute the speed, we merge the 3-channel results by the same strategy to plot the Fig. \ref{fig3} and we also get their best binary results to measure the four metrics. Our method achieves the comparable performance with the best method in blindness detection and outperform all other methods by a large margin in terms of speed. The performance of our model only trained on our KITTICS dataset also proves that our synthetic dataset is authentic. Fig. \ref{fig4} shows the visual comparison of all methods, our method can handle hard cases and accurate. Fig. \ref{fig5} shows some of our results on our KITTICS test dataset. Our network correctly predicts the blindness types and accurately estimates the blindness maps. 

\subsection{Evaluation on the KITTICS Dataset}
We also conduct the comparison of our method and other existing methods on our KITTICS Dataset. As Table \ref{table5} shows, our approach achieves high accuracy and speed.

\begin{table}[h]
	\vspace{0.2cm}
	\begin{center}
		\newcommand{\tabincell}[2]{\begin{tabular}{@{}#1@{}}#2\end{tabular}}
		\resizebox{86mm}{10mm}{
			\begin{tabular}{|c|c|c|c|c|c|c|}
				\hline
				& \multicolumn{4}{|c|}{Motion Blur and Defocus Blur}&\multicolumn{2}{|c|}{Haze}\\
				\hline
				& Shi \textit{et al.} \cite{shi2014discriminative} & Tang \textit{et al.} \cite{tang2016spectral}&\tabincell{c}{Golestaneh \\ \textit{et al.}} \cite{golestaneh2017spatially}&  Ours & He \textit{et al.} \cite{he2010single} & Ours\\
				\hline
				MAE$\downarrow$ & 0.371 & 0.507 & 0.169 & \textbf{0.048} & 0.109 & \textbf{0.047}\\
				\hline
				MSE$\downarrow$ & 0.254 & 0.351& 0.058 & \textbf{0.010} & 0.031 & \textbf{0.004}\\
				\hline
				Spped$\uparrow$ & 0.004 & 0.472 & 0.007 & \textbf{227.894} & 3.935 & \textbf{239.120}\\
				\hline
			\end{tabular}
		}
		\caption{Evaluation of blindness map estimation on our KITTICS test dataset in terms of mean-squared-error (MSE) and mean-absolute-error (MAE).}
		\label{table4}
	\end{center}
	\vspace{-0.6cm}
\end{table}

\subsection{Ablation Study}

We do ablation study to find out the effect of each part of our network on blindness map estimation and blindness type classification performance. 

We compare four different network architectures. Comparing to the final architecture, the first removes the blur maps from the input for  \textbf{\textit{BTC}} (w/o-bm). The second removes the auxiliary mask (w/o-mask). The third removes both of them. The compared performance from the final architecture is shown in Table \ref{table5}.   

As shown in Table \ref{table5}, our final model outperforms other network architectures. And the experiment also proves both the two operations can improve the accuracy of blindness maps. All these models are able to achieve high classification accuracy because of the parallel CNN layers. This strategy broadens the network and reduce the influence of inter-class correlation.  

\begin{table}[h]
	\begin{center}
		\newcommand{\tabincell}[2]{\begin{tabular}{@{}#1@{}}#2\end{tabular}}
		\resizebox{86mm}{10mm}{
			\begin{tabular}{|c|c|c|c|c|}
				\hline
				& W/o-bm-mask & W/o-bm & W/o-mask & Final\\
				\hline
				MAE$\downarrow$ & 0.041 & 0.037 & 0.037
				 & \textbf{0.036} \\
				\hline
				MSE$\downarrow$ & 0.007 & 0.006 & 0.006 & \textbf{0.006} \\
				\hline
				\tabincell{c}{Classification \\ accuracy }$\uparrow$ & 1.0& 1.0& 1.0& \textbf{1.0}\\
				\hline 
			\end{tabular}
		}
		\caption{Ablation study of blindness map estimation and blindness type classification on our KITTICS test dataset. We use the results of classification to choose which channel of output blur maps to evaluate.}
		\label{table5}
	\end{center}
	\vspace{-0.4cm}
\end{table}

\subsection{Generalization Ability}
\begin{figure}[H]
	\vspace{-0.4cm}
	\centering
	\includegraphics[scale=0.32]{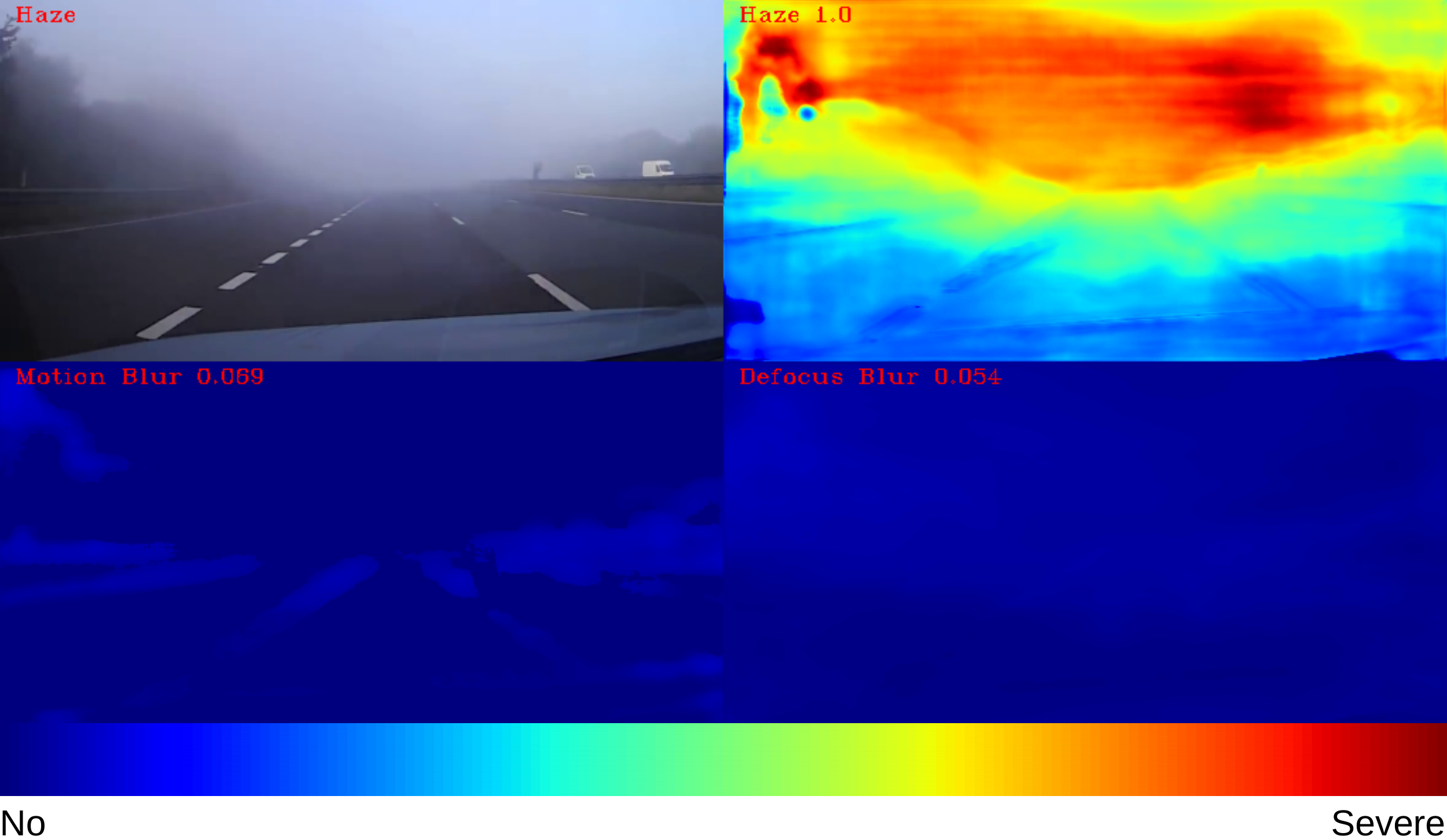}
	\caption{Our results on real driving scene. From top to bottom and left to right: the input image, the output haze map, the output motion blur map, the output defocus blur map and the color bar. The red texts represent the blindness types and their possibilities.}
	\label{fig6}
	\vspace{-0.2cm}
\end{figure}

We collect some real scenes videos from the Internet to test the generalization ability of our method and the results are shown on our supplementary video. Our model trained on our dataset achieves convincing results in Fig. \ref{fig6}. Our network exactly identifies the scene as haze and estimates the hazy degree of each pixel in the output haze map. Furthermore, our model also detects the motion blur and the defocus blur which are slight. 

\section{CONCLUSIONS}
In this paper, we focused on the safety problem of outdoor robots and autonomous driving in particular. We designed a blindness quantification method, built a synthetic dataset named KITTICS and proposed a lightweight network architecture for simultaneous blindness map estimation and blindness type classification. Despite annotation by labelers, our quantification method automatically provided robust ground truth values. Our dataset has a larger number of images and more blindness types than existing datasets. It could also be used for other tasks, such as deblurring or dehazing. The experimental results demonstrated our network architecture could achieve accurate and reliable performance in short timescales. The ability of our method for generalization is also proven. Thus, our approach could be widely used for other outdoor robotics systems, which often encounter extreme blindness conditions and need to run online. Therefore, our approach is a great help for safe decision making in robotic systems. 
%\addtolength{\textheight}{-12cm}

%%%%%%%%%%%%%%%%%%%%%%%%%%%%%%%%%%%%%%%%%%%%%%%%%%%%%%%%%%%%%%%%%%%%%%%%%%%%%%%%

%%%%%%%%%%%%%%%%%%%%%%%%%%%%%%%%%%%%%%%%%%%%%%%%%%%%%%%%%%%%%%%%%%%%%%%%%%%%%%%%

%%%%%%%%%%%%%%%%%%%%%%%%%%%%%%%%%%%%%%%%%%%%%%%%%%%%%%%%%%%%%%%%%%%%%%%%%%%%%%%%
%\section*{ACKNOWLEDGMENT}

%%%%%%%%%%%%%%%%%%%%%%%%%%%%%%%%%%%%%%%%%%%%%%%%%%%%%%%%%%%%%%%%%%%%%%%%%%%%%%%%

\bibliographystyle{IEEEtran}
\bibliography{DeepBlur}

\end{document}